\documentclass[sigconf,nonacm]{acmart}

\AtBeginDocument{%
  \ifxetex
    \setmathfont{latinmodern-math.otf}%
  \fi
}

\setcopyright{none}
\acmConference[KDD Cup 2026 UniRec Workshop]
{KDD Cup 2026 Tencent UniRec Challenge Workshop}
{August 12, 2026}
{Jeju, Korea}
\acmYear{2026}
\acmDOI{}
\acmISBN{}

\usepackage{amsmath,amsfonts}
\usepackage{graphicx}
\usepackage{booktabs}
\usepackage{hyperref}
\usepackage{xcolor}
\usepackage{multirow}
\usepackage{CJKutf8}

\newcommand{\cn}[1]{\begin{CJK}{UTF8}{gbsn}#1\end{CJK}}

\hypersetup{
  colorlinks=true,
  linkcolor=blue,
  citecolor=blue,
  urlcolor=blue,
}

\title{QueryFormer: Winning Solution for KDD Cup 2026 Tencent UniRec Challenge}

\author{Yuanzhe Zhou}
\affiliation{%
  \institution{Wuhan University}
  \city{}
  \country{}
}
\email{}

\author{Zhaoyang Zeng}
\affiliation{%
  \institution{Sun Yat-sen University}
  \city{}
  \country{}
}
\email{}

\renewcommand{\shortauthors}{Y. Zhou et al.}

\begin{document}

\begin{abstract}
Post-click conversion rate (pCVR) prediction requires jointly modeling feature interactions and sequential user behaviors. The KDD Cup 2026 Tencent UniRec Challenge calls for a unified architecture addressing both. We observe that existing unified architectures often generate query tokens---the central information hub---with projection-based multi-layer perceptrons (MLPs), without explicit token-to-query attention for refining the query side. We propose \textbf{QueryFormer}, centered on a \textbf{stackable unified field--sequence block} that bridges non-sequential multi-field features and behavioral sequences, and provide a latency-aware scaling study over view width $H$, model width, depth, data, and compute. The block generates queries through cross-attention and packs sequence queries into shared-parameter attention. QueryFormer secured \textbf{1st place} in the Industrial Track, achieving an official test area under the ROC curve (AUC) of 0.83254; a modest post-competition scale-up reached 0.832713. Within our grid, $H$-scaling improves validation AUC from 0.84540 to 0.84615 and beats HyFormer at comparable budgets. Ablation identifies query generation as the largest contributor. Packed shared-parameter cross-attention keeps $H{=}8$ inference latency to only $1.89{\times}$ that of $H{=}1$, positioning the bridge as an efficient stackable unified block.
\end{abstract}

\maketitle

\begin{figure*}[t]
\centering
\includegraphics[width=0.96\textwidth]{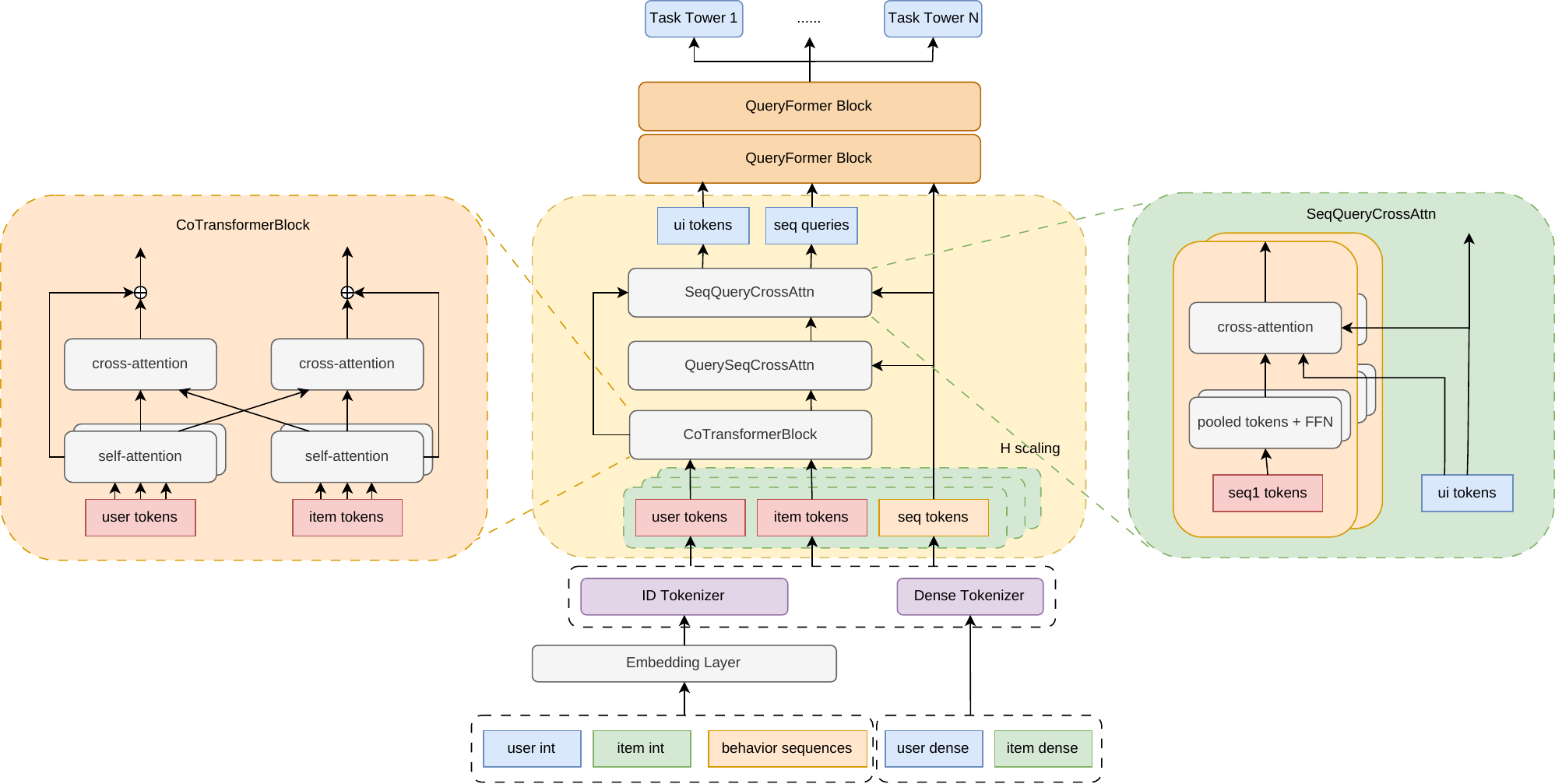}
\caption{QueryFormer architecture. The reusable stackable query-generation bridge takes non-sequential field tokens and behavioral sequence tokens as input, returns generated query probes and mixed representations, and preserves the $H$ dimension as parallel tokenization columns with shape-$(B_s,H,T,d_{\text{model}})$ views, where $B_s$ is batch size and $T$ is token count, before per-column outputs are combined.}
\Description{Architecture diagram showing QueryFormer's query-driven bridge from non-sequential tokens to behavioral sequences through CoTransformerBlock, QuerySeqCrossAttn, and SeqQueryCrossAttn, with H independent tokenization columns.}
\label{fig:architecture}
\end{figure*}

\begin{figure*}[tb]
\centering
\includegraphics[width=0.93\textwidth]{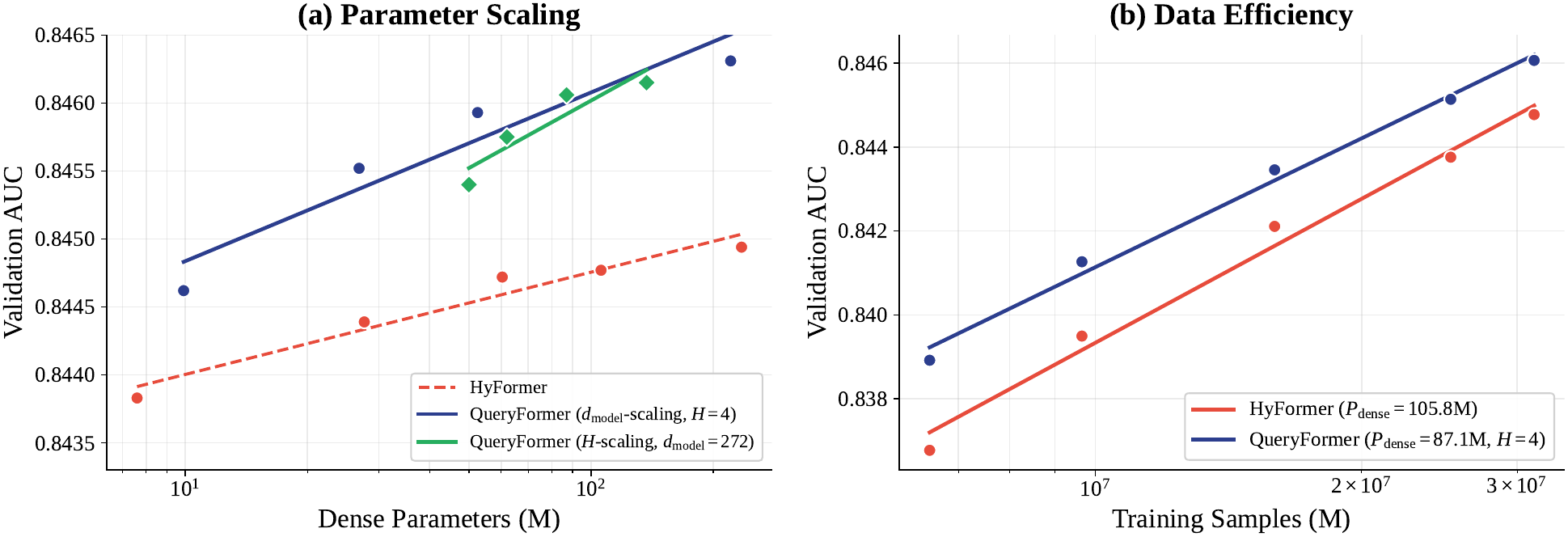}
\caption{Scaling evidence for the stackable query-generation bridge. \textbf{(a)} Parameter scaling: QueryFormer traces a stronger AUC--dense-parameter frontier than HyFormer~\cite{huang2026hyformer} within the explored budgets, with separate $d_{\text{model}}$- and $H$-scaling paths. \textbf{(b)} Data efficiency: QueryFormer ($P_{\mathrm{dense}}{=}87.1\mathrm{M}$, $H{=}4$) remains above HyFormer ($P_{\mathrm{dense}}{=}105.8\mathrm{M}$) from about 20\% to full training data. Both x-axes are logarithmic; $P_{\mathrm{dense}}$ denotes dense parameter count.}
\Description{Two-panel scaling analysis. (a) Parameter scaling: AUC vs dense parameters for QueryFormer d\_model-scaling, H-scaling, and HyFormer. (b) Data efficiency: AUC vs training samples for HyFormer (P\_dense=105.8M) and QueryFormer (P\_dense=87.1M, H=4). Both subplots use log x-axes.}
\label{fig:scaling_comparison}
\end{figure*}

\section{Introduction}

The KDD Cup 2026 Tencent UniRec Challenge requires post-click conversion rate (pCVR) models that jointly handle multi-field features and user behavior sequences. Our solution, \textbf{QueryFormer}, won the Industrial Track. The key observation is that existing unified architectures often generate query tokens with projection-based multi-layer perceptrons (MLPs), which provide nonlinear context compression but lack explicit token-to-query attention. Since the field-to-sequence cross-attention channel $\langle N,B\rangle$ (non-sequential field tokens attending to behavioral sequence tokens) is dominant~\cite{liu2026est}, query quality directly determines what sequence evidence is retrieved and how user-item context is aggregated.

QueryFormer redesigns query processing as a stronger query-generation mechanism. It combines token-level self/cross attention, item-to-sequence retrieval, and cross-attention-based query generation (Figure~\ref{fig:architecture}). Its reusable unit is a stackable unified field--sequence block: field and behavior tokens enter a token-in/token-out bridge, generated query probes retrieve sequence evidence, and packed shared-parameter attention keeps multi-view execution efficient. Ablations show that replacing SeqQueryCrossAttn with an MLP causes the largest degradation, and scaling the embedding matrix width $H$ gives consistent gains. Figure~\ref{fig:scaling_comparison} previews the scaling results against HyFormer.

Our key contributions are:
\begin{enumerate}
    \item \textbf{Query Generation Mechanism}: cross-attention-driven query generation adds explicit token-to-query attention beyond projection-based MLP generation.
    \item \textbf{Unified Block}: a token-in/token-out field--sequence bridge composes with token-based recommenders and uses packed shared-parameter attention.
    \item \textbf{Latency-Aware Scaling Study}: $H=4$ independent tokenization columns form a shape-$(B_s,H,T,d_{\text{model}})$ embedding matrix, enabling batch-parallel scaling across views, depth, width, data, and compute.
    \item \textbf{Industrial Validation}: 1st place in the KDD Cup 2026 Industrial Track (test AUC 0.83254) with consistent scaling trends across multiple dimensions.
\end{enumerate}

\section{Related Work}

\subsection{Feature Interaction and Sequence Modeling}
DeepFM~\cite{guo2017deepfm}, DCN-V2~\cite{wang2021dcnv2}, Fi-GNN~\cite{li2019fignn}, and DLRM~\cite{naumov2019dlrm} model static feature interactions through product, cross-network, graph, or embedding-interaction layers. Sequential recommenders such as DIN~\cite{zhou2018din}, DIEN~\cite{zhou2019dien}, SIM~\cite{pi2020sim}, BERT4Rec~\cite{sun2019bert4rec}, SASRec~\cite{kang2018selfattentive}, and HSTU~\cite{zhai2024hstu} extract interest from behavior histories. These lines are usually optimized as separate modules; QueryFormer instead lets field tokens and behavior evidence interact through a unified query pipeline.

\subsection{Token-Based Architectures}
Recent token-based models~\cite{ding2023hyperformer,huang2026hyformer,zhang2025onetrans,tokenformer2026} represent heterogeneous recommendation features as tokens processed by Transformer-style blocks. EST~\cite{liu2026est} shows that $\langle N,B\rangle$ cross-attention is the dominant interaction channel, but it does not optimize how the query side is generated. HyFormer and OneTrans reuse MLP-generated queries; LENS, GAP-Net, and HeMix calibrate MLP queries with gates, cascades, or mixed static/dynamic components. QueryFormer takes a stronger step: it generates query probes through cross-attention and repeatedly refines them through a query-centric pipeline (Table~\ref{tab:positioning}). In Table~\ref{tab:positioning}, Bridge describes how explicitly a method connects non-sequential field tokens with behavioral sequence evidence during query generation.

\begin{table}[!htbp]
\centering
\caption{Unified-block positioning. Tok$\to$Q: explicit token-to-query attention; token I/O: token input/output; Packed: shared-parameter packed attention for multi-view queries.}
\label{tab:positioning}
\footnotesize
\setlength{\tabcolsep}{2.8pt}
\begin{tabular}{@{}l c c c c c@{}}
\toprule
\textbf{Method} & \textbf{Query} & \textbf{Tok$\to$Q} & \textbf{Bridge} & \textbf{Views} & \textbf{Packed} \\
\midrule
HyFormer~\cite{huang2026hyformer} & MLP & $\times$ & implicit & $\times$ & $\times$ \\
HeMix~\cite{wang2026hemix} & Mixed & $\times$ & mixed & $\times$ & $\times$ \\
LENS~\cite{wang2026lens} & Gate & $\times$ & gated & $\times$ & $\times$ \\
GAP-Net~\cite{wu2026gapnet} & Cascade & $\times$ & cascade & $\times$ & $\times$ \\
\midrule
\textbf{QueryFormer} & \textbf{Cross-Attn} & \textbf{\checkmark} & \textbf{token I/O} & \textbf{$H$} & \textbf{\checkmark} \\
\bottomrule
\end{tabular}
\end{table}

Recent scaling studies in recommendation~\cite{li2026expand} motivate treating capacity allocation as a first-class design axis. For optimization, we combine Muon-style dense updates~\cite{jordan2024muon,kaggle2025cmi,kaggle2026nfl}, Adagrad for sparse embeddings~\cite{duchi2011adagrad}, and exponential moving average (EMA) weight averaging~\cite{izmailov2018averaging}.

\section{Preliminaries}

Unified token-based architectures embed heterogeneous features into a shared $d_{\text{model}}$-dimensional space. We denote non-sequential (NS) user tokens as $N_u$, item tokens as $N_i$, their union as $N=N_u\cup N_i$, and behavioral sequence tokens from $S$ domains as $B=\{B_1,\ldots,B_S\}$. For two token sets $X$ and $Y$, $\langle X,Y\rangle$ denotes a cross-attention interaction using $X$ as queries and $Y$ as keys/values. We use this notation below to distinguish item-to-sequence retrieval from behavior-initialized query generation.

Existing architectures often generate queries through projection-based MLPs:
\begin{equation}
    \mathbf{q} = \text{MLP}\big(\text{pool}(B) \oplus \mathbf{N}\big)
\end{equation}
where $\oplus$ denotes concatenation. Such MLPs are nonlinear and can fuse pooled context, but they do not explicitly let query tokens attend to field or sequence tokens. Recent methods calibrate query quality via gating~\cite{wang2026lens} or cascading~\cite{wu2026gapnet}; QueryFormer instead generates queries through context-aware attention.

\section{Methodology}

\subsection{Overview}

QueryFormer processes user/item sparse features, dense features, four behavior domains, and temporal context through a unified token pipeline. Its two tiers are \textbf{Query-Driven Architecture} (Section~\ref{sec:query}), where three attention stages build high-quality query representations, and \textbf{Supporting Components} (Section~\ref{sec:supporting}), which provide tokenization and interaction infrastructure.

\subsection{Query-Driven Architecture}
\label{sec:query}

\noindent\fbox{\parbox{\dimexpr\columnwidth-2\fboxsep-2\fboxrule\relax}{
\textbf{Design Principles.}
\begin{enumerate}
\item \textbf{Generate through attention, not projection alone.} Cross-attention over NS tokens yields data-dependent query probes beyond projection-only summaries.
\item \textbf{Refine through interaction.} Queries are enriched via self-attention and cross-attention before they retrieve behavioral evidence.
\item \textbf{Scale through diversity.} The embedding matrix provides $H$ independent tokenization views, improving the explored scaling frontier.
\end{enumerate}
}}

\subsubsection{Query-Generation Pipeline}

The query-generation pipeline updates the NS context and then constructs final query probes:
\begin{align}
N' &= \mathrm{FeatureInteract}(N_u,N_i), \\
R &= \mathrm{SeqRetrieve}(N'_i,B), \\
Q &= \mathrm{QueryGenerate}(\mathrm{pool}(B),N'\cup R), \\
Y &= \mathrm{Mix}(Q,N',B).
\end{align}
Here $N'_i$ denotes the item-token slice of $N'$. This pipeline exposes the reusable bridge component of QueryFormer: NS tokens first retrieve sequence evidence, then generated query tokens re-query the refined NS context. The bridge targets the field--sequence interaction point of token-based recommenders: it takes field tokens and behavior tokens as input, and returns generated query probes plus mixed representations for downstream towers. Because its interface is token-in/token-out, the bridge is architecturally stackable with existing unified recommenders rather than tied to a single backbone. It is also computationally stackable: sequence queries are concatenated and evaluated by shared-parameter cross-attention, amortizing key/value projections and graphics processing unit (GPU) kernel launches instead of invoking cross-attention serially. Table~\ref{tab:modules} gives the operator-to-mechanism mapping.

Table~\ref{tab:modules} lists the four attention mechanisms organized into three architectural stages. Together they form a \emph{query-centric} refinement pipeline: queries are enriched via self-attention, cross-interact with counterpart tokens, attend to behavioral sequences, and drive query generation through explicit token-to-query attention.

\begin{table}[t]
\centering
\caption{Attention mechanisms by architectural stage. For SeqQueryCrossAttn, $Q_B$ denotes behavior-initialized query probes.}
\label{tab:modules}
\footnotesize
\begin{tabular}{@{}l l l l@{}}
\toprule
\textbf{Stage} & \textbf{Mechanism} & \textbf{Type} & \textbf{Interaction} \\
\midrule
\multirow{2}{*}{Co-Transformer} & QuerySelfAttn & SelfAttn & $\langle N_u,N_u\rangle$, $\langle N_i,N_i\rangle$ \\
                               & QueryCrossAttn & CrossAttn & $\langle N_u,N_i\rangle$ \\
\midrule
QuerySeqCrossAttn & --- & CrossAttn & $\langle N_i,B\rangle$ \\
SeqQueryCrossAttn & --- & CrossAttn & $\langle Q_B,N\rangle$ \\
\bottomrule
\end{tabular}
\end{table}

\subsubsection{Embedding Matrix}

\begin{figure}[t]
\centering
\includegraphics[width=0.92\columnwidth]{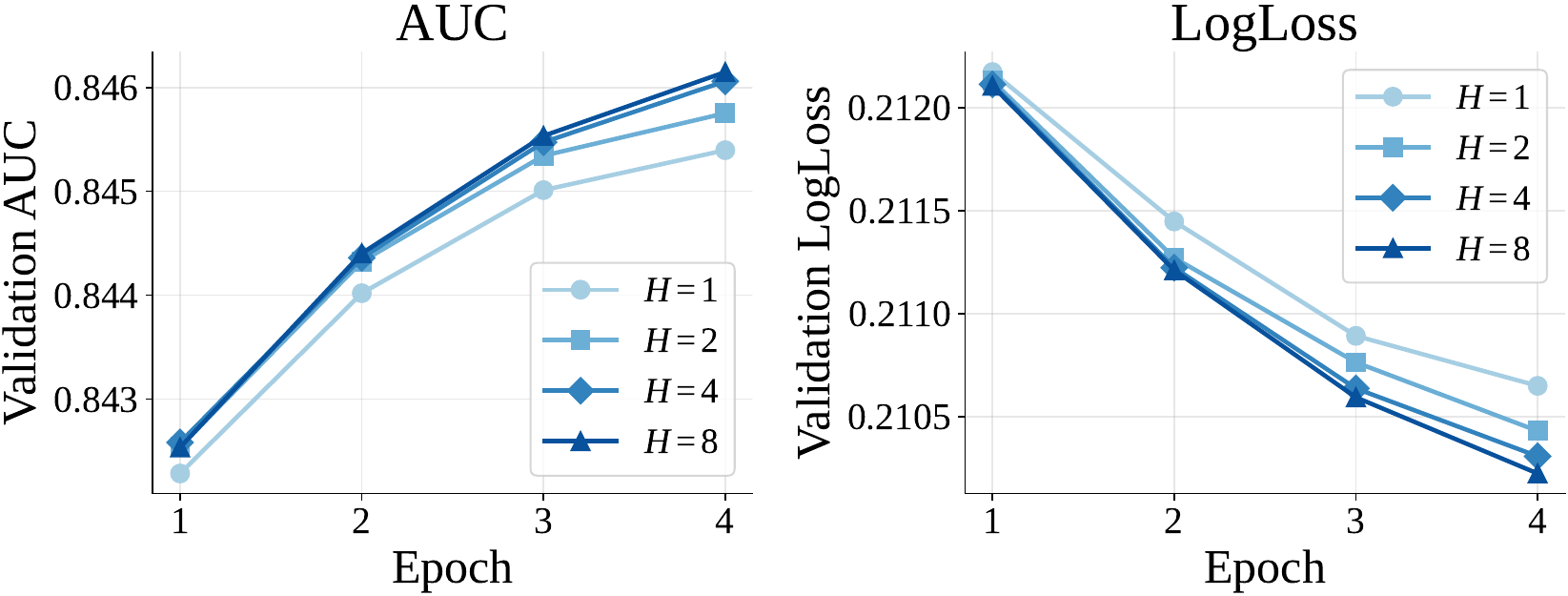}
\caption{Validation AUC and LogLoss vs.~training steps for $H\in\{1,2,4,8\}$ at $d_{\text{model}}{=}272$.}
\Description{Training curves showing that wider embedding matrices achieve better final AUC and LogLoss.}
\label{fig:scaling_h}
\end{figure}

Instead of a flat token sequence, QueryFormer uses an embedding matrix with $H$ independent columns. Each column tokenizes the same input through its own pipeline, producing a tensor of shape $(B_s,H,T,d_{\text{model}})$, where $B_s$ is batch size and $T$ is token count. This implements view diversity in one batch-parallel forward pass and exposes a direct scaling axis. Figure~\ref{fig:scaling_h} shows that increasing $H$ from 1 to 8 consistently improves validation AUC and LogLoss at fixed $d_{\text{model}}=272$.

\subsubsection{Co-Transformer}

The Co-Transformer contains two layers. Each layer first applies QuerySelfAttn within user and item token sets, then QueryCrossAttn between user and item tokens with gated fusion. This separates within-field refinement from cross-side interaction and lets us ablate the two effects independently.

\subsubsection{Retrieval}

QuerySeqCrossAttn retrieves behavior evidence by using aggregated item NS tokens as queries over each of the four behavior domains. The retrieved vectors summarize item-relevant historical evidence and are appended to the NS token set.

\subsubsection{SeqQueryCrossAttn}

SeqQueryCrossAttn generates $N_q=2$ query probes per sequence domain. Mean/max-pooled behavior summaries initialize probes, and cross-attention over refined NS tokens makes them context-aware:
\begin{equation}
    \mathbf{q}_{d,k}=\mathrm{CrossAttn}_d(\mathbf{q}_{d,k}^{(0)},\mathbf{N},\mathbf{N}).
\end{equation}

\subsection{Supporting Components}
\label{sec:supporting}

\subsubsection{NS Token Construction}
NS tokens are built from user/item integer features, dense feature slices, temporal features, item ID, and four retrieved sequence tokens, yielding 26 NS tokens.

\subsubsection{Dense Feature Fusion}
User and item dense features are processed by DenseFusionModel, which combines a low-rank DCN-V2 layer ($L=1$, rank $r=8$), a SiLU-activated MLP, and Squeeze-and-Excitation (SENet) recalibration~\cite{hu2018senet}.

\subsubsection{Sequence Token Embedding}
Each behavior domain embeds side-information independently, projects it to $d_{\text{model}}$, and adds a 65-bucket recency embedding.

\subsubsection{Blocks}
Following HyFormer and OneTrans~\cite{huang2026hyformer,zhang2025onetrans}, $K=2$ lightweight processing blocks evolve behavior sequences with Swish-Gated Linear Unit (SwiGLU) encoders, attend from queries to sequences, and mix tokens through parameter-free permutation.

\subsection{Training Recipe}

\subsubsection{Multi-Loss Supervision}

The training objective combines binary cross-entropy (BCE) with an auxiliary mean absolute error (MAE) loss on log-transformed conversion time ($\lambda=0.1$). Removing it reduces validation AUC from 0.84606 to 0.84590, a smaller effect than any attention ablation.

\subsubsection{Optimization}

We use \textbf{MuonPlus} for dense parameters, \textbf{Adagrad} for sparse embeddings, CosineAnnealingLR, gradient clipping at $\ell_2$-norm 1.0, EMA decay 0.999, \texttt{bfloat16}, and \texttt{torch.compile}.

\section{Experiments}

\subsection{Experimental Setup}

Input data is stored in Apache Parquet format with a 90/10 Row-Group-level train/validation split across 142 feature columns and four behavioral domains (a, b, c, d). All experiments use a fixed sequence limit $L_{\mathrm{seq}}=256$: sequences are padded to 256 positions and longer histories are truncated to the most recent positions. Labels include conversion type and log-transformed conversion timestamp.

We evaluate on the Round 2 Industrial Track dataset using area under the ROC curve (AUC) as the primary metric, coupled with LogLoss (logarithmic loss) and inference latency constraints. In this leaderboard setting, AUC differences around $10^{-4}$ were ranking-relevant; the top two official submissions differ by 0.00037. QueryFormer targets the competition's innovation criteria with a stackable field--sequence bridge and a latency-aware scaling study over $H$, $K$, $d_{\text{model}}$, training length, data, and compute cost. Evidence is organized as follows: Table~\ref{tab:baseline_comparison} compares the optimized recipe with HyFormer, Table~\ref{tab:ablation} isolates the bridge mechanisms, Table~\ref{tab:h_efficiency} measures packed-attention latency, and Table~\ref{tab:scaling}/Figure~\ref{fig:scaling_comparison} summarize multi-axis scaling.

Validation AUC is consistently higher than test AUC (best val 0.84631 vs.\ test 0.83254), reflecting the expected distribution shift between temporally proximate train/val splits and the held-out test period. The relative ordering was preserved for submitted architectural and scaling changes, so we report validation AUC for most experiments due to limited test submission slots. Models are trained on NVIDIA GPUs with multi-GPU distributed data parallel (DDP) via \texttt{torchrun}. The PyTorch implementation uses a custom \texttt{IterableDataset} with pre-allocated NumPy buffers, row-group-balanced DDP sharding, EMA-shadow validation, and sidecar configs for checkpoint recovery. For final submission, training on the full 35M samples (no validation holdout) yields $\sim$$+$0.0001 AUC, and ensemble distillation with multiple variants supervising a single student~\cite{tang2018ranking,ding2026recdistill,lai2025exploring,zhu2020ensembled} yields $\sim$$+$0.0007 test AUC at single-model inference cost. Due to the competition timeline, we did not exhaustively sweep all high-capacity configurations; the post-competition result suggests additional gains remain available.

\begin{table}[H]
\centering
\caption{Validation comparison under the optimized training recipe. Dense parameters are in millions; all runs use $L_{\mathrm{seq}}{=}256$.}
\label{tab:baseline_comparison}
\small
\begin{tabular}{@{}l r r r r@{}}
\toprule
Model & $H$ & Dense & Val AUC & LogLoss \\
\midrule
HyFormer~\cite{huang2026hyformer} & -- & 105.8 & 0.84477 & -- \\
QueryFormer & 4 & 87.1 & 0.84606 & 0.210309 \\
QueryFormer & 4 & 220.8 & 0.84631 & 0.210241 \\
\bottomrule
\end{tabular}
\end{table}

\subsection{Ablation Study}
\label{sec:ablation}

\begin{figure}[t]
\centering
\includegraphics[width=0.92\columnwidth]{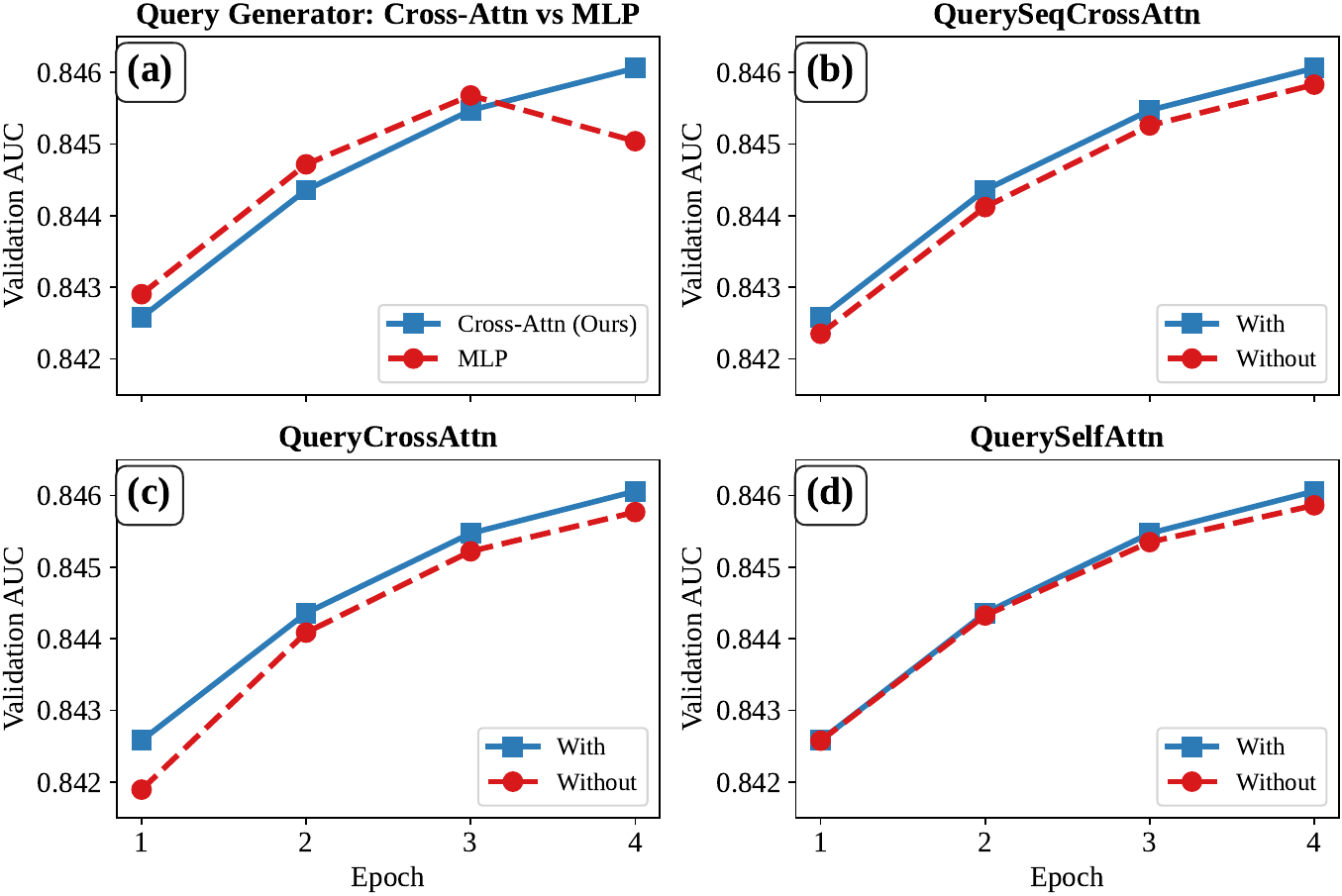}
\caption{Per-epoch ablation of four attention mechanisms.}
\Description{Validation AUC across training epochs for four ablation configurations: replacing cross-attention query generation with MLP causes the largest drop, followed by removing QueryCrossAttn, QuerySeqCrossAttn, and QuerySelfAttn, confirming each mechanism independently contributes.}
\label{fig:query_ablation}
\end{figure}

Table~\ref{tab:ablation} reports ablation results; Figure~\ref{fig:query_ablation} shows per-epoch trends. The degradation order---cross-attention query generation causes the largest drop, followed by cross-attention, sequence attention, and self-attention---confirms that query quality is the primary performance driver. Removing all four mechanisms compounds the loss ($\Delta=-0.00110$), validating the additive benefit of each component.

\begin{table}[t]
\centering
\caption{Ablation of attention mechanisms ($H{=}4$, $d_{\text{model}}{=}272$).}
\label{tab:ablation}
\small
\begin{tabular}{@{}l r r@{}}
\toprule
Configuration & Best AUC & $\Delta$ \\
\midrule
Full Model (baseline) & 0.84606 & --- \\
~~- Replace SeqQueryCrossAttn with MLP & 0.84568 & $-$0.00038 \\
~~- Remove QueryCrossAttn & 0.84577 & $-$0.00029 \\
~~- Remove QuerySeqCrossAttn & 0.84583 & $-$0.00023 \\
~~- Remove QuerySelfAttn & 0.84586 & $-$0.00020 \\
\midrule
~~- Remove all four attention mechanisms & 0.84496 & $-$0.00110 \\
\bottomrule
\multicolumn{3}{@{}r@{}}{\footnotesize All runs: $K{=}2$, 4 epochs.}
\end{tabular}
\end{table}

\subsection{Competition Results}

Table~\ref{tab:leaderboard} shows the final leaderboard of the Industrial Track. QueryFormer ranks 1st with a test AUC of 0.83254. Post-competition, a modest parameter scale-up improved test AUC to 0.832713; due to limited post-competition time, this configuration was not exhaustively tuned and likely remains below the best attainable setting.

\begin{table}[H]
\centering
\caption{Industrial Track leaderboard (top 3), with post-competition improvement.}
\label{tab:leaderboard}
\small
\begin{tabular}{@{}c l r r@{}}
\toprule
Rank & Team & Test AUC & $\Delta$ \\
\midrule
1st & \textbf{QueryFormer} (\cn{日之光面}) & \textbf{0.83254} & --- \\
2nd & RClaw & 0.83217 & $-$0.00037 \\
3rd & load\_state\_dict & 0.83145 & $-$0.00109 \\
\midrule
-- & \textbf{QF-scaled-post} & \textbf{0.832713} & $+$0.000173 \\
\bottomrule
\end{tabular}
\end{table}

\subsection{Scaling Analysis}

Model capacity was a binding constraint in this competition: the 35M-sample dataset demanded representations beyond a single-column architecture. The scaling study asks which stackable axis gives the best AUC-latency trade-off: view width $H$, block depth $K$, model width $d_{\text{model}}$, or training compute. We report packed-attention latency separately because Round 2 imposed tight inference limits.

\begin{table}[H]
\centering
\footnotesize
\setlength{\tabcolsep}{3.2pt}
\renewcommand{\arraystretch}{1.04}
\caption{Latency-aware $H$ scaling of the packed bridge on 31.35M training samples and 3.48M validation samples.}
\label{tab:h_efficiency}
\begin{tabular}{@{}c c c c c@{}}
\toprule
$H$ & Train Time & Eval Time & Eval Throughput & Latency / sample \\
\midrule
1 & 43.7 min & 1.8 min & 32.2k samples/s & 31.1 $\mu$s \\
2 & 56.0 min & 2.0 min & 29.0k samples/s & 34.5 $\mu$s \\
4 & 80.1 min & 2.5 min & 23.2k samples/s & 43.1 $\mu$s \\
8 & 137.3 min & 3.4 min & 17.0k samples/s & 58.7 $\mu$s \\
\bottomrule
\end{tabular}
\end{table}

\begin{table}[H]
\centering
\footnotesize
\setlength{\tabcolsep}{3.1pt}
\renewcommand{\arraystretch}{0.98}
\caption{Multi-axis scaling grid (SwiGLU encoders; params in millions). Unspecified values use $K{=}2$, $H{=}4$, $d_{\text{model}}{=}272$, and 4 epochs.}
\label{tab:scaling}
\begin{tabular}{@{}l r r r r@{}}
\toprule
Setting & Sparse & Dense & Val AUC & LogLoss \\
\midrule
$H{=}1$ & 530 & 50 & 0.84540 & 0.210651 \\
$H{=}2$ & 563 & 62 & 0.84575 & 0.210432 \\
$H{=}4$ & \textbf{628} & \textbf{87} & \textbf{0.84606} & \textbf{0.210309} \\
$H{=}8$ & 757 & 137 & 0.84615 & 0.210225 \\
\midrule
$K{=}1$ & 628 & 77 & 0.84598 & 0.210302 \\
$K{=}3$ & 628 & 97 & 0.84616 & 0.210270 \\
\midrule
11 epochs & 628 & 87 & 0.84616 & 0.210712 \\
\midrule
$d{=}68$ & 557 & 10 & 0.84462 & 0.210942 \\
$d{=}136$ & 581 & 27 & 0.84552 & 0.210540 \\
$d{=}204$ & 604 & 53 & 0.84593 & 0.210360 \\
$d{=}544$ & 722 & 221 & 0.84631 & 0.210241 \\
\bottomrule
\end{tabular}
\end{table}

Table~\ref{tab:scaling} presents a scaling analysis across four dimensions within our explored grid. First, increasing $H$ from 1 to 8 monotonically improves AUC and LogLoss, making view diversity the strongest observed scaling direction. Under Round-2's tight latency constraint, Table~\ref{tab:h_efficiency} identifies $H{=}4$ as the practical operating point for a GPU-efficient unified architecture: it gains $+$0.00066 AUC over $H{=}1$ with validation time increasing from 1.8 to 2.5 minutes. The packed shared-attention implementation keeps latency growth sublinear: $H{=}8$ is only $1.89{\times}$ slower than $H{=}1$ (58.7 vs. 31.1 $\mu$s/sample), rather than approaching an $8{\times}$ serial cost. Second, increasing the block count from $K{=}1$ to $K{=}3$ gives a smaller gain at fixed $H{=}4$ and $d_{\text{model}}{=}272$, suggesting that depth-only stacking is not the dominant path in our grid. Third, extending training from 4 to 11 epochs gives only $+$0.00010 AUC and worse LogLoss, suggesting compute saturation. Fourth, increasing $d_{\text{model}}$ improves AUC but with diminishing returns as dense parameters grow quadratically. Together, these results support multi-axis stackability: the bridge scales across $H$ views, depth, width, data/compute, and packed execution, rather than only repeated layers.

\subsubsection{Architectural Scaling Efficiency}

Figure~\ref{fig:scaling_comparison} compares QueryFormer and HyFormer~\cite{huang2026hyformer} under the optimized recipe (MuonPlus, EMA, Adagrad, SwiGLU). QueryFormer achieves higher AUC at comparable dense-parameter budgets, while data scaling shows the same trend from about 20\% to full data. Fitting $\Delta\mathrm{AUC}=C\cdot P^k$ following UniMixer~\cite{huang2026unimixer}, where $P$ denotes dense parameter count and QF/HF abbreviate QueryFormer/HyFormer, gives:
\begin{align}
\Delta\mathrm{AUC}_{\text{QF},d} &= 3.82{\times}10^{-4} \cdot P^{0.286} \quad (R^2{=}0.896), \\
\Delta\mathrm{AUC}_{\text{QF},H} &= 8.7{\times}10^{-6} \cdot P^{0.924} \quad (R^2{=}0.811), \\
\Delta\mathrm{AUC}_{\text{HF}} &= 1.9{\times}10^{-4} \cdot P^{0.326}.
\end{align}
Within the explored range, $H$-scaling is close to linear while $d_{\text{model}}$-scaling saturates. We view these fits as empirical scaling diagnostics rather than universal laws. They suggest a practical rule for this dataset: allocate budget to view diversity before representation depth; broader high-capacity sweeps remain future work.

\section{Conclusion}

QueryFormer improves unified recommendation by generating query tokens through explicit token-to-query attention rather than projection-only MLP generation. As a unified block, its reusable contribution is a token-in/token-out field--sequence bridge with efficient packed attention. As a scaling study, it identifies view width $H$ as a strong axis alongside depth, width, data, and compute. The evidence is consistent across three paths: cross-attention query generation is the strongest ablated component, $H$ exposes a stronger scaling axis than standard width in our grid, and packed execution keeps $H{=}8$ latency to only $1.89{\times}$ that of $H{=}1$. QueryFormer won the KDD Cup 2026 Industrial Track, with post-competition results indicating further headroom under larger sweeps. Limitations include reliance on one large-scale dataset and the latency cost of wider embedding matrices.

\setlength{\bibsep}{0pt}
\renewcommand{\bibfont}{\footnotesize}
\bibliographystyle{ACM-Reference-Format}
\bibliography{references}

\end{document}